\documentclass[10pt,twocolumn,letterpaper]{article}

\usepackage{wacv}
\usepackage{times}
\usepackage{epsfig}
\usepackage{graphicx}
\usepackage{amsmath}
\usepackage{amssymb}

\usepackage{times}
\usepackage{epsfig}
\usepackage[linesnumbered, ruled]{algorithm2e}
\usepackage{multirow}
\usepackage{rotating}
\usepackage{amsmath}
\usepackage{mathtools}
\usepackage{makecell}
\usepackage{booktabs}
\usepackage{dblfloatfix}

\usepackage{footnote}
\usepackage[bottom]{footmisc}
\usepackage{lipsum}
\graphicspath{{figs/}}

\wacvfinalcopy 

\def\wacvPaperID{0528} 

\ifwacvfinal\fi
\begin{document}

\title{Advanced Super-Resolution using Lossless Pooling Convolutional Networks}

\author{Farzad Toutounchi, Ebroul Izquierdo \\
Multimedia and Vision Research Group\\Queen Mary University of London\\
{\tt\small \{f.toutounchi,e.izquierdo\}@qmul.ac.uk}
}

\maketitle
\ifwacvfinal\thispagestyle{empty}\fi

\begin{abstract}
In this paper, we present a novel deep learning-based approach for still image
super-resolution, that unlike the mainstream models does not rely solely on the 
input low resolution image for high quality upsampling, and takes advantage
of a set of artificially created auxiliary self-replicas of the input image that
are incorporated in the neural network to create an enhanced and accurate 
upscaling scheme. Inclusion of the proposed lossless pooling layers, and the fusion
of the input self-replicas enable the model to exploit the high correlation
between multiple instances of the same content, and eventually result in significant
improvements in the quality of the super-resolution, which is confirmed
by extensive evaluations.

\end{abstract}

\section{Introduction}
\label{sec:intro}
Super-resolution (SR) has received significant attention in recent years, and the learning-based
approaches in designing SR solutions have been providing promising results for spatial
upsampling of still images and videos. In particular, application of deep learning and 
Convolutional Neural Networks (CNN) has become a trendy approach for SR, leading to
major improvements in the performance of the state-of-the-art SR solutions.

Dong et al.~\cite{dong2014,dong2016a} presented the first
concrete deep learning-based SR approach for still images, exploiting deep CNNs, capable 
of upsampling still images with high visual quality and reasonable complexity. 
Dong et al.~improved their work further in~\cite{dong2016b} 
by introduction of transposed convolutional layers to their framework for upsampling 
from source to target resolution.
Shi et al.~\cite{shi2016} used a similar concept as~\cite{dong2016b} for reducing the 
complexity of the framework, except they introduced a sub-pixel upscaling 
layer instead of the transposed convolution layer, with which 
they achieved an efficient performance.

Other major contributions in still image SR using deep learning have 
been made by Kim et al.~\cite{kim2016a,kim2016b}. In~\cite{kim2016a}, the concept of 
residual learning was introduced to deep SR models, which 
led to quicker convergence of the network and high visual quality reconstruction. 
In~\cite{kim2016b}, a deep and recursive architecture was employed, that achieved
very good performance in terms of image quality, although expensive in terms of computation 
cost. In other deep learning-related efforts, Wang et al.~\cite{wang2015} introduced a deep joint SR model to 
exploit both external and self-similarities for reconstruction, in which a stacked denoising 
convolutional auto-encoder was first trained on external training data, then it was
fine-tuned with multi-scale self-examples from the input data.

\begin{figure}[t!]
\begin{center}
\fontsize{7pt}{7pt}\selectfont
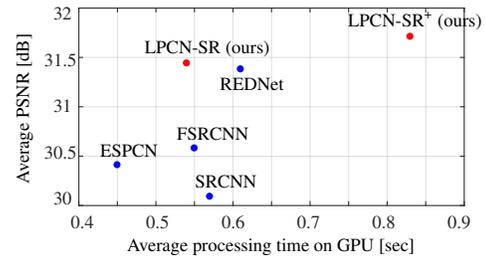
\vspace{-5pt}
\end{center}
\caption{Performance comparison of the proposed SR approach with the state-of-the-art models on Set 5 images.}
\label{fig:set5}
\end{figure}

Another interesting endeavor in devising deep learning models for image SR 
was done by Mao et al.~\cite{mao2016}. They introduced a deep hourglass-shaped CNN 
that included multiple convolution and transposed convolution layers with symmetric skip connections and was 
designed originally for image denoising. However, the model was also applied for SR
and image upscaling, and it was reported as one of the best SR models in the literature for 
high quality image upscaling. Ledig et al.~\cite{ledig2017} also made a contribution to SR by using a generative adversarial
network for performing high quality SR on images, and they showed promising results in
highly textured images, and high subjective quality, although the method falls behind the state-of-the-art approaches
in terms of the objective metrics.

\begin{figure*}[b!]
\begin{center}
\fontsize{7pt}{7pt}\selectfont
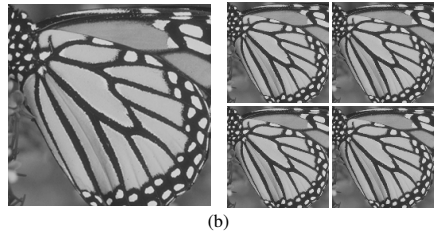
\vspace{-5pt}
\end{center}
\caption{The lossless pooling process: (a) Input and output of the layer for $r=2$, and (b) a sample lossless pooling on a grayscale image.}
\label{fig:superpixel}
\end{figure*}

While major improvements are reported in the still image SR domain, video SR has also 
been an active research field, and several interesting contributions have been made
in recent years by applying deep learning-based solutions to multi-frame SR \cite{caballero2017,kappeler2016,tao2017,makansi2017}.
In multi-frame SR, unlike the typical still image SR problem, several input frames that are
highly correlated contribute in generation of one high quality upsampled target image, hence
a better objective and subjective quality can be expected when compared to the single image SR.
Inspired by this approach, we aimed at devising a model that utilizes the multi-frame concept
within the still image frameworks for an enhanced SR experience, using artificially generated
self-replicas of the input image.

In this paper, we propose a novel approach for still image SR which exploits the incorporation of 
several auxiliary downscaled versions of the input image in CNNs, that results in a significant 
improvement in the quality of the reconstructed output image (Figure~\ref{fig:set5}). The followings summarize the main contributions 
of this paper:
\begin{itemize}
    \item Introduction of lossless pooling layers that create several highly correlated downsampled replicas
from the input image, that can fuse in the SR network. 
    \item Designing a lossless pooling convolutional network SR (LPCN-SR) model, that only exploits the
downsampled self-replicas, and can provide faster performance than the existing high quality approaches.
    \item Designing a lossless pooling convolutional network SR model that also incorporates the original input
image in addition to the self-replicas (LPCN-SR\textsuperscript{+}), and can outperform the existing approaches 
in terms of quality.
\end{itemize}

The rest of this paper is organized as follows. In Section~\ref{proposed}, we introduce the novelties
including the lossless pooling mechanism, as well as incorporation of the self-replicas in SR networks. 
Section~\ref{architecture} covers the technical specifications of the proposed CNN architecture and the 
proposed SR model. Section \ref{experiments} summarizes the results of our evaluations and comparisons 
with state-of-the-art approaches, followed by conclusions in Section~\ref{conc}.

\section{Proposed Approach}
\label{proposed}
Our proposed approach is inspired by the multi-frame SR models that take several highly 
correlated low resolution inputs and generate a high resolution version of a target image. In
order to extend this concept to still image SR, we require multiple versions of the low resolution
image as the input. Hence a mechanism is needed to create different replicas of the input 
image. We propose a particular pooling process which downsamples the input image to several lower
resolution versions by reshuffling the pixel positions and without any information loss. The resulting
replicas of the input image can be incorporated in a multi-frame model for upsampling the target image.
In the following, the proposed pooling operation is described in details. The process of self-replicas
fusion and their inclusion in the SR network are described next.

\subsection{Lossless Pooling Layer}
\label{sec:superpixel}

Normal pooling layers in CNNs result in the loss of data, however we propose
a pooling layer that downscales a single-channel image to a multi-channel image with lower spatial resolution.
Lossless pooling process is performed by rearranging a $H \times W$ matrix $\mathbf{M}$ into
a $\frac{H}{r} \times \frac{W}{r} \times r^2$ tensor $\mathbf{T}$. This operation can be considered
as a reverse sub-pixel upscaling defined in~\cite{shi2016}, and can be described mathematically
as the following $\mathcal{LP}$ function:
\begin{equation}
\label{eq:1}
\mathbf{T} = {\mathcal{LP}(\mathbf{M})}_{x,\ y,\ c} = \mathbf{M}_{r\cdot x -\bmod(r^2 - c, r),\ r\cdot y - \left \lfloor \frac{r^2-c}{r} \right \rfloor}
\end{equation}
where x, y, and c represent the coordinates of a pixel in the output tensor. Although the
above description aims at lossless pooling of single-channel (grayscale) images, the
concept can be easily extended to multi-channel images. Figure~\ref{fig:superpixel}
demonstrates the lossless pooling operation, along with a visual example.

Application of lossless pooling layer reduces the spatial size of the input data without
losing any information, and reduces the computation cost of convolution operation by a factor
of $r^2$. Moreover, the process results in several replicas of the input image
with very high correlations, which can eventually act as a data augmentation process that can 
enhance the still image SR.

\subsection{Self-Replicas Fusion}
\label{multi}

\begin{figure*}[t!]
\begin{center}
\fontsize{7pt}{7pt}\selectfont
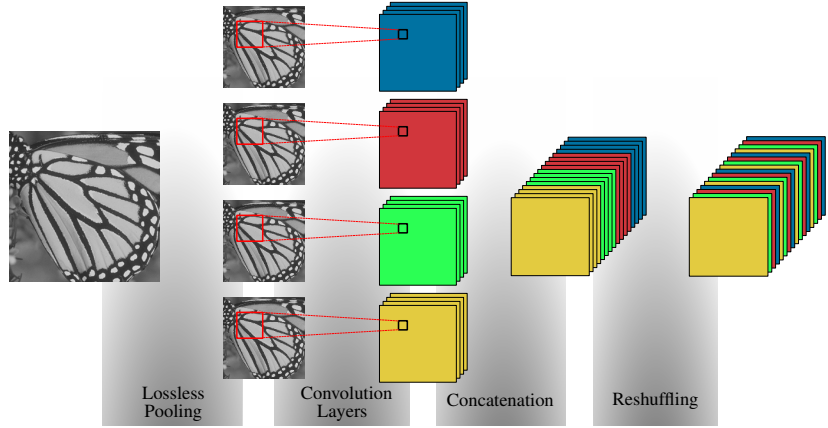
\vspace{-5pt}
\end{center}
\caption{The process of fusing the self-replicas in CNNs by concatenation and reshuffling of the convolution outputs.}
\label{fig:fusion}
\end{figure*}

In multi-frame SR, application of multiple input frames for creating a high quality upsampled target frame is well
studied, and Caballero et al.~\cite{caballero2017} and Kapperler et al.~\cite{kappeler2016} present different ways of
fusing the inputs within the CNN architectures to obtain a single-image output from a multi-frame input, while fully
exploiting the correlations between the input frames. We adopt a similar approach, namely early fusion, for coping
with the downsampled replicas created by the lossless pooling layer.

If an input frame of size $H \times W$ is fed to the lossless pooling layer with parameter $r$, $r^2$ 
downsampled replicas are generated, and each replica is fed to a convolutional layer with $n$ filters. The resulting
output is an ensemble of $r^2$ feature maps, each with the size $(\frac{H}{r}, \frac{W}{r}, n)$. The feature
maps can then be concatenated and create one set of feature maps $\mathbf{F}$ with the size $(\frac{H}{r}, \frac{W}{r}, n \times r^2)$.
This is similar to the early fusion concept introduced in \cite{caballero2017,kappeler2016}, also demonstrated
in Figure~\ref{fig:fusion} for the case of $r=2$ and $n=4$.

In addition to the concatenation of the feature maps, which is a widely used approach in multi-frame processing, we 
perform a reshuffling of the feature maps in order to create a better mix of the features produced by different replicas.
The reshuffling is performed by rearranging the order of the features in the depth dimension as the following $\mathcal{RS}$
function:
\begin{equation}
\label{eq:reshuffling}
{\mathcal{RS}(\mathbf{F})}_{x,\ y,\ c} = {\mathbf{F}}_{x,\ y,\ \left \lceil \frac{c}{r^2} \right \rceil + n \cdot \bmod(c-1, r^2)} 
\end{equation}
where the reshuffled output of the process is depicted in Figure~\ref{fig:fusion}. This output can be treated
as a normal feature map and be further processed in a CNN with different layers. Similar to the lossless pooling
process, the concatenation and reshuffling processes can also be easily extended to multi-channel images.

\section{Architecture and Implementation}
\label{architecture}

\begin{figure*}[t!]
\begin{center}
\fontsize{7pt}{7pt}\selectfont
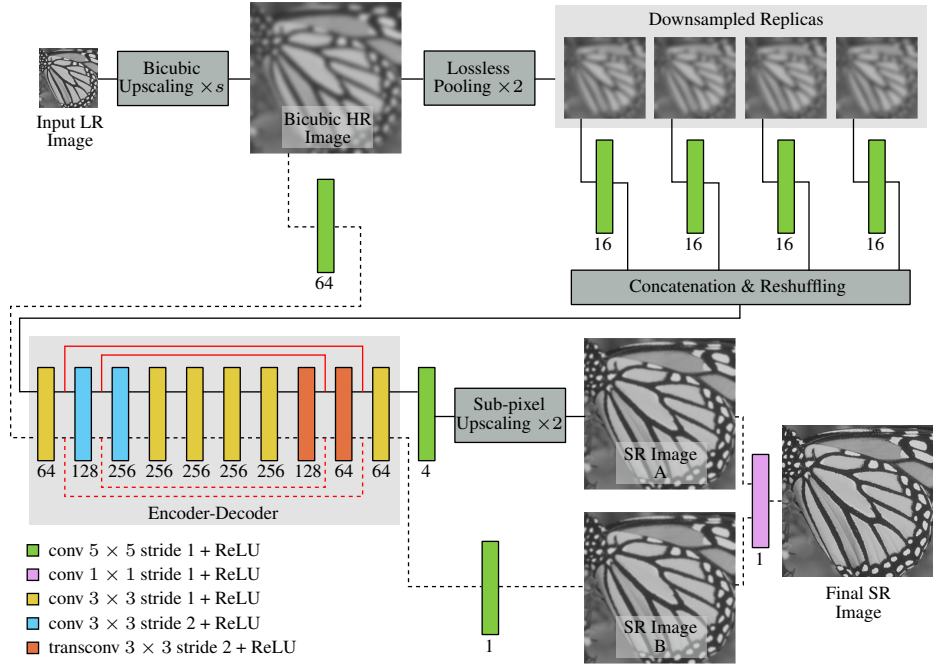
\vspace{-5pt}
\end{center}
\caption{The architecture of the proposed CNN for still image SR. The solid connections depict the LPCN-SR model. Inclusion of
the dashed connections in the model results in the LPCN-SR\textsuperscript{+}. The red lines present the skip connections between layers.}
\label{fig:arch}
\end{figure*}

With the introduction of the lossless pooling and the concatenation and reshuffling mechanisms, a CNN architecture can 
be devised to incorporate the abovementioned processes in the still image SR problem. Our proposed network is illustrated 
in Figure~\ref{fig:arch}. The first step in the process is a bicubic interpolation similar to many state-of-the-art approaches
to reach the target resolution. This step also ensures that one architecture can be applied for upsampling with different 
scaling factors. Hence given an input low resolution image, a high resolution version using bicubic upscaling
is created, and then the lossless pooling, individual convolutional layers on downsampled replicas, concatenation of the feature maps, 
and reshuffling is performed. Depending on the selected architecture, the network can work on two different modes, each described
in details in the following.

\subsection{LPCN-SR}
\label{only}
The proposed CNN architecture for still image SR can be operational in two modes. The first mode is a basic version of the model,
namely LPCN-SR, that only takes the downsampled replicas as the key inputs for SR, and ignores the original low quality input image generated by bicubic 
interpolation. This model is depicted by solid lines and connections in Figure~\ref{fig:arch}. The output to this model is denoted as
\emph{SR Image A} in the figure. As mentioned earlier, an input grayscale image of size $(\frac{H}{s}, \frac{W}{s})$ is interpolated to 
the target resolution of $(H, W)$ using bicubic filtering. The first step after the bicubic interpolation is the lossless pooling 
with $r=2$, which results in four downsampled replicas, each with the size of $(\frac{H}{2}, \frac{W}{2})$. Each of the replicas goes through
separate convolutional layers with 16 filters, resulting in four tensors of size $(\frac{H}{2}, \frac{W}{2}, 16)$. The four tensors are concatenated 
and reshuffled according to Section~\ref{multi} to create a feature map of size $(\frac{H}{2}, \frac{W}{2}, 64)$.  

The resulting feature map is then fed to a typical encoder-decoder architecture consisting of 10 layers with the kernel specifications presented
in Figure~\ref{fig:arch}. The encoder-decoder structure contains coupled convolution and transposed convolution layers with stride of 2, 
that are also connected using skip connections. The output of this stage is a tensor with a similar size as the input feature map to 
the encoder-decoder network. The generated feature map is then fed to one last convolution layer resulting in a tensor of size
$(\frac{H}{2}, \frac{W}{2}, 4)$. We then use a sub-pixel upscaling operation as proposed in~\cite{shi2016} to upsample the tensor to 
the target resolution and create an image of the target size $(H, W)$. This image is denoted as \emph{SR Image A}, and is a high quality
reconstruction of the input image created by the downsampled replicas as the only input to the model.

\subsection{LPCN-SR\textsuperscript{+}}
\label{total}
In addition to the downsampled replicas created by the lossless pooling, the original bicubic version of the image can also be incorporated in the network,
and provide more information for creating a higher quality SR image. This process is depicted using the dashed lines and connections in 
Figure~\ref{fig:arch}, complementing the model described previously. The first step after the bicubic filtering is a convolution layer
with 64 filters that results in a feature map of size $(H, W, 64)$. This feature map is then fed to the same encoder-decoder network described
earlier with the same weights. We apply weight sharing for this section of the model in order to avoid extra complexity and having too many 
network parameters. The output of the encoder-decoder network will be a feature map of size $(H, W, 64)$, which is then fed to a single kernel
convolution layer, that results in a high quality SR image, denoted as \emph{SR Image B} in the figure. 

The two created SR images, \emph{SR Image A} and \emph{SR Image B}, are then combined using a single kernel convolution layer with the kernel
size of $1 \times 1$ to create the \emph{Final SR Image}. The final convolution layer operates as an averaging mechanism to mix the two created
SR images. It is worth noting, that the dashed section of the model in Figure~\ref{fig:arch} is essentially a similar concept to the REDNet model 
proposed in \cite{mao2016}.

\subsection{Training}
\label{train}
Training the proposed SR model is performed by solving an optimization
problem to minimize the error between the ground truth labels and the network outputs. 
The training for the basic and the full models can be performed separately. 
For both cases, the training labels are a set of high resolution image 
samples $\mathbf{X}$, and the network outputs are the high quality high 
resolution images $\mathbf{X}^\ast$ generated by the model (\emph{SR Image A} for LPCN-SR and 
\emph{Final SR Image} for LPCN-SR\textsuperscript{+}). The network outputs can
be formulated as the following:
\begin{equation}
\label{eq:model}
\mathbf{X}^\ast = {\boldsymbol{\theta}}_m(\mathbf{Y})
\end{equation}
where $\mathbf{Y}$ is the input low resolution image, $\boldsymbol{\theta}$ is the end-to-end SR model
encompassing all the parameters, and $m$ defines the network model of operation with is either LPCN-SR or
LPCN-SR\textsuperscript{+}.

\begin{table*}[!b]
  \caption{Quality and complexity analysis of the proposed method and state-of-the-art approaches for the scaling factor of 4.}
  \vspace*{7pt}
\fontsize{9.25pt}{9.25pt}\selectfont
    \centering
    \begin{tabular}{ccccccccc}
    \toprule[1pt]
        &  & \bf{Bicubic} & \bf{SRCNN} & \bf{ESPCN} & \bf{FSRCNN}  & \bf{REDNet}  & \bf{LPCN-SR} & \bf{LPCN-SR\textsuperscript{+}}\\
        \midrule[1pt]
\multirow{4}{*}{\begin{turn}{90}\bf{Set 5}\end{turn}} & PSNR  &  28.44 & 30.09 & 30.41 & 30.58 & 31.38 & 31.44 & \bf{31.71} \\
                                                     & SSIM  &  0.8110 & 0.8520 & 0.8590 & 0.8658 & 0.8820 & 0.8833 & \bf{0.8872} \\
                                                     & GPU time & - & 0.57 & 0.45 & 0.55 & 0.61 & 0.54 & 0.83 \\  
                                                     & CPU time & - & 0.11 & 0.02 & 0.02 & 0.27 & 0.27 & 1.17 \\
        \midrule[1pt]
\multirow{4}{*}{\begin{turn}{90}\bf{Set 14}\end{turn}} & PSNR  &  26.00 & 27.18 & 27.37 & 27.52 & 27.98 & 27.98 & \bf{28.12} \\
                                                      & SSIM  & 0.7009 & 0.7385 & 0.7457 & 0.7506 & 0.7636 & 0.7643 & \bf{0.7686} \\
                                                      & GPU time & - & 0.60 & 0.46 & 0.58 & 0.70 & 0.56 & 1.03 \\
                                                      & CPU time & - & 0.22 & 0.03 & 0.03 & 0.55 & 0.53 & 2.39 \\
        \midrule[1pt]
\multirow{4}{*}{\begin{turn}{90}\bf{BSD 100}\end{turn}} & PSNR  &  25.89 & 26.64 & 26.77 & 26.85 & 27.16 & 27.14 & \bf{27.26} \\
                                                    & SSIM  & 0.6651 & 0.6994 & 0.7073 & 0.7100 & 0.7207 & 0.7215 & \bf{0.7253} \\ 
                                                    & GPU time & - & 0.57 & 0.44 & 0.57 & 0.68 & 0.59 & 0.96 \\
                                                    & CPU time & - & 0.15 & 0.02 & 0.02 & 0.36 & 0.36 & 1.57 \\
        \bottomrule[1pt]
     \end{tabular}
    \label{tab:results}
\end{table*}

The mean squared error, defined as the following, is employed as the cost function for the training process.
\begin{equation}
\label{eq:mse}
J_{MSE}(\mathbf{X}, \mathbf{Y}, \boldsymbol{\theta}_m) = \frac{1}{N} \sum_{i=0}^{N}\left ( {\mathbf{X}}_i - {\boldsymbol{\theta}}_m ({\mathbf{Y}}_i) \right )^2
\end{equation}
with $N$ denoting the total number of training samples.
The training input samples are created by downsampling the high resolution samples,
and upscaling them back to the original resolution by bicubic interpolation. The scaling
factor can be fixed to focus on training a particular scaling, or alternatively can include
several values to cover a wide range of scaling.

\section{Experiments}
\label{experiments}
We focused the experiments on the SR with a scaling factor of 4, which is a challenging factor
in image scaling, and it is also the basis for most of the SR evaluations. We used the DIV2K data 
set~\cite{div2k} as our training set, which comprises 800 high quality images. The images were 
partitioned into $96 \times 96$ samples with a stride of 80, which led to around 325,000 training 
sample pairs. We used Adam optimizer~\cite{adam} for training the model with a learning rate of 
0.0001, $\beta_1$ of 0.9, $\beta_2$ of 0.999, $\epsilon$ of $10^{-8}$ and a training batch
size of 128. The proposed model, along with the existing state-of-the-art approaches, were implemented
using TensorFlow\footnote{https://www.tensorflow.org/} library.

\subsection{Quantitative Evaluations}
\label{sec:comparison}
To compare the presented model with the existing solutions, we used Set 5, Set 14 and BSD 100 data sets, which
are widely used in SR task. We compared the approach with well-known deep learning-based
SR models including SRCNN~\cite{dong2014},
ESPCN~\cite{shi2016}, FSRCNN~\cite{dong2016b}, and REDNet~\cite{mao2016}.
All the models were implemented and trained according to the provided information in the literature. All the trainings
were performed on Tesla K80 NVIDIA GPUs, and all the tests were performed on a machine with a generic
Intel Core i7-6700 CPU with a 3.40GHz clock and 16GB RAM, and a GeForce GTX 1070 GPU.

Table~\ref{tab:results} summarizes the performance results of the presented model, in comparison with
state-of-the-art methods. The quality metrics are the Peak Signal-to-Noise Ratio (PSNR) and Structural 
SIMilarity (SSIM) index. According to the results, the proposed approach can outperform the baseline models
in all datasets, promising a high quality enhancement for still images.

The basic LPCN-SR model can perform as good as the best existing model, REDNet, and in some cases outperform
it slightly. The strength of this model however is in the lower computation cost on GPU, which is due to using
lossless pooling, that results in downscaling of the input, and consequently performing all the convolutional
processes in a lower resolution than the native target resolution. The full LPCN-SR\textsuperscript{+} model,
on the other hand, shows a solid outperformance on all data sets and provides major improvements in PSNR and
SSIM results. The computation cost is however higher than the existing models due to the structure of the model, 
and incorporating multiple input signals in the approach.

\begin{figure*}[t!]
\begin{center}
\fontsize{7pt}{7pt}\selectfont
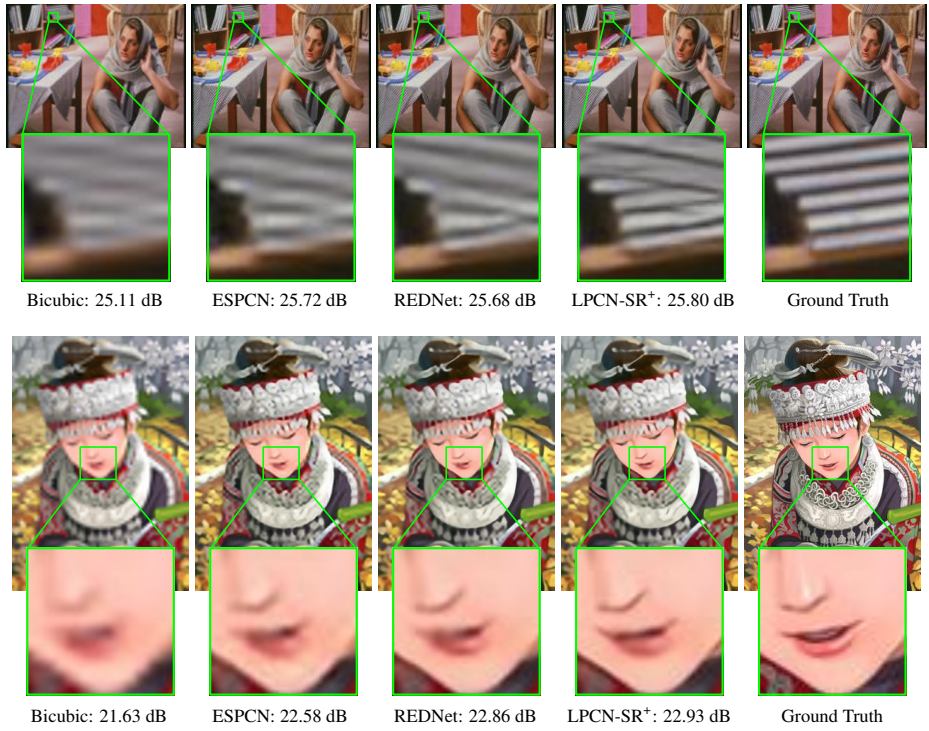
\vspace{-5pt}
\end{center}
\caption{Upscaling \emph{Barbara} and \emph{Comic} from Set 14 with a factor of 4.}
\label{fig:subj}
\end{figure*}

\begin{figure*}[t!]
\begin{center}
\fontsize{7pt}{7pt}\selectfont
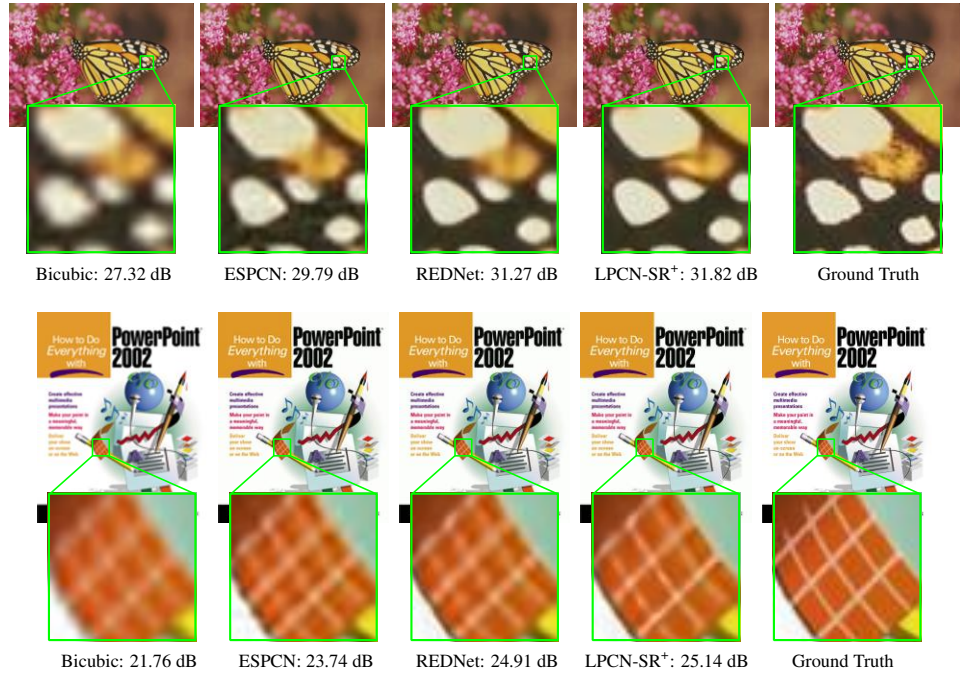
\vspace{-5pt}
\end{center}
\caption{Upscaling \emph{Monarch} and \emph{PPT3} from Set 14 with a factor of 4.}
\label{fig:subj_2}
\end{figure*}

\subsection{Qualitative Evaluations}
\label{sec:subjective}
We also examined the subjective quality of the enhanced images on the test content, as objective metrics cannot always
grasp the intricacies detected by human eyes. Application of the proposed approach resulted in clear visual 
improvements in the reconstructed high resolution images, some of which are presented in Figure~\ref{fig:subj} and Figure~\ref{fig:subj_2}.
In these examples only the luma signal is upsampled by the proposed SR approach, and the color components
are scaled using bicubic interpolation. 

One of the most challenging scenarios in SR is coping with the low resolution images with major aliasing 
distortions. The \emph{Barbara} image from Set 14 of the test sets is an extreme example for this case. Upscaling 
the downsampled version of this image can result in magnifying those aliasing artifacts in most cases. However, as
depicted in Figure~\ref{fig:subj}, our proposed model does a good job in restoring some of the content that is
highly distorted by the downsampling process, and is outperforming the other SR approaches. Another example is
in the \emph{PPT3} image, depicted in Figure~\ref{fig:subj_2}, which reconstruction of a distorted patterned 
texture is shown, and LPCN-SR\textsuperscript{+} provides a better prediction of the original image compared to 
the state-of-the-art models.

\section{Conclusions}
\label{conc}
Still image SR can benefit from application of auxiliary correlated inputs derived from the 
original low resolution source input. The process of creating self-replicas and their incorporation in 
the CNNs is performed by a novel lossless pooling layer, that generates multiple downscaled
versions of the input image, which are later fused in the general SR framework. The presented
LPCN-SR model can perform as good as the state of the art with a reduced computation cost on 
GPUs, and the full LPCN-SR\textsuperscript{+} outperforms the existing approaches, and promises
high quality and accurate image upscaling.

\section*{Acknowledgments}
The work described in this paper has been conducted within the project
COGNITUS. This project has received funding from the European Union’s
Horizon 2020 research and innovation program under grant agreement No
687605. This research utilized Queen Mary's Apocrita HPC facility,
supported by QMUL Research-IT. The authors gratefully acknowledge the support of NVIDIA 
Corporation with the donation of the Titan X Pascal GPU used for this research.

{\small
\bibliographystyle{ieee}
\bibliography{references}
}

\end{document}